\documentclass[runningheads, a4paper]{llncs}

\usepackage{geometry}
\usepackage{multirow}
\usepackage{graphicx}
\begin{document}
\title{Active learning for object detection in high-resolution satellite images}
\titlerunning{Active learning for object detection}
%
\author{Alex Goupilleau \and
Tugdual Ceillier \and
Marie-Caroline Corbineau}
\authorrunning{A. Goupilleau et al.}
%
\institute{Preligens (ex-Earthcube), Paris, France \\ \url{www.preligens.com}, [name].[surname]@preligens.com}
\maketitle              
\begin{abstract}
In machine learning, the term \emph{active learning} regroups techniques that aim at selecting the most useful data to label from a large pool of unlabelled examples. While supervised deep learning techniques have shown to be increasingly efficient on many applications, they require a huge number of labelled examples to reach operational performances. Therefore, the labelling effort linked to the creation of the datasets required is also increasing. When working on defense-related remote sensing applications, labelling can be challenging due to the large areas covered and often requires military experts who are rare and whose time is primarily dedicated to operational needs. Limiting the labelling effort is thus of utmost importance. This study aims at reviewing the most relevant active learning techniques to be used for object detection on very high resolution imagery and shows an example of the value of such techniques on a relevant operational use case: aircraft detection.

\keywords{Deep learning  \and Convolutional neural networks \and Earth observation \and Active learning \and Object detection.}
\end{abstract}
\section{Introduction}

\subsection{Context}

Active learning is a sub-domain of machine learning, whose objective is to smartly select the relevant data to label in order to maximize the performance of a given model. Active learning can help achieve two goals:
\begin{itemize}
\item Get the best possible performance given a fixed labelling budget, 
\item Minimize the labelling effort to reach some target performance.  
\end{itemize}

In remote sensing use cases, the labelling effort is particularly laborious, as one often has to find objects of interest in very large images, before creating a label for them. The integration of active learning methods could therefore allow annotators to focus only on the most relevant images or zones inside images -- which would make the whole labelling process easier, faster, and more efficient.

In particular, deep learning models -- including convolutional neural networks (CNN) -- would greatly benefit from these methods, as these types of machine learning models generally need a large amount of labelled images to work well. In a larger context, active learning methods could help improve the whole lifecycle of an algorithm: on a regular basis, some raw images could be ingested in a database of unlabelled images, of which only a relevant subset would be selected by active learning methods in order to be labelled and used for the improvement of a given deep learning model.

These techniques are particularly interesting when the user has limited labelling capabilities and owns a lot of unlabelled images -- and one can expect the latter to be especially true in the future where the number of sensors is assumed to increase over time.

\subsection{Problem setting}

The objective of this section is to formulate the problem that active learning methods try to solve, as well as to establish the necessary vocabulary to understand the different methods that will be described in the following subsections.   

The classic active learning scenario can be described in the following way: a given operator owns a database of unlabelled images $B$ and a CNN classification model $M$ -- that takes as input an image $X$ and outputs class probabilities $p(X\in c) = M(X)$.   

The operator has a given labelling budget: this means that he can only label a maximum of $N$ images in the database $B$. After selection and labelling, the resulting labelled dataset can be called $B*$. The objective of any active learning method is to find the optimal $B*$ subset of images in $B$ that maximizes the performances of the model $M$ on its task, when trained on $B*$. 

\begin{figure}[bt]
    \centering
    \includegraphics[width=12cm]{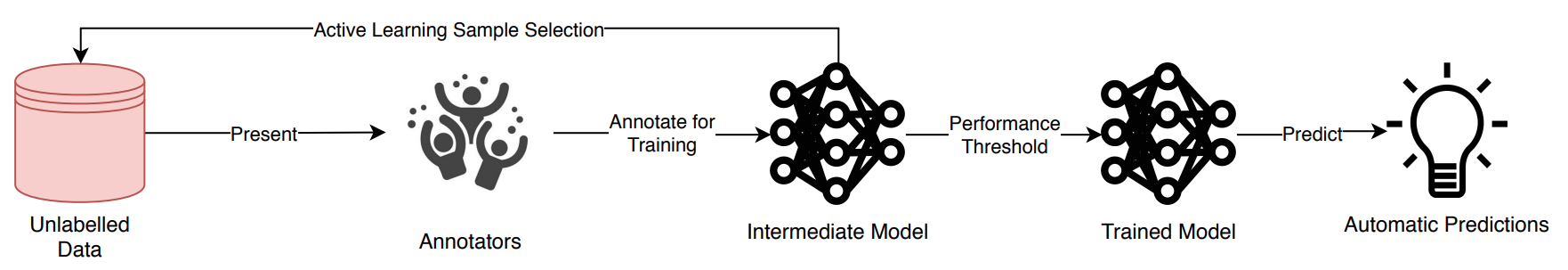}
    \caption{\label{fig:active_pipeline}%
    \textbf{Schematics of the active learning iterative process (from \cite{budd_survey_2019}).}}
\end{figure}

In practice, this scenario is often iterative. The operator already owns a dataset of labelled images that was used to train an initial model. Then, at each step,  the active learning technique selects new unlabelled images that are then labelled and used to retrain or fine-tune the initial model, either on the concatenation of the initial dataset and the new images or on the selected images only. This iterative process can be stopped if targeted performances or maximum labelling budget are reached, or continuously used to accommodate for novelties in the data distribution (in the remote sensing use case, it can be new types of objects, new geographic areas, new sensors). Figure~\ref{fig:active_pipeline} illustrates this iterative process.

\section{Related work}

The objective of an active learning method is to evaluate the informativeness of a given image $X$, i.e. how much information that image can bring to the model $M$ when it is trained on it. To do so, most techniques use the model itself. The different approaches used by active learning methods differ in the way informativeness is defined. The different strategies can be split into four categories: uncertainty-based active learning, representativity-based active learning, hybrid methods that combine uncertainty and representativity, and methods that use neither. In this section, we will detail examples of these four categories.

\subsection{Uncertainty-based active learning}

The techniques based on uncertainty select images that generate the most uncertainty for the model.  A well known and relevant technique to estimate model uncertainty is the one used by \cite{gal_deep_2017}. In order to evaluate uncertainty, the authors use Bayesian dropout at inference time to obtain slightly different predictions. The authors test different ways of selecting images based on these multiple predictions, called \emph{acquisition functions}. Using the MNIST dataset, they show that compared to a random selection of the images, all acquisition functions yield positive performance gains -- except one based on the variance -- and that these techniques work especially well if the labelling budget is low.

Other approaches also use uncertainty in order to select relevant images to label, either by using an ensemble of classifiers \cite{beluch_power_2018}, or by using generative adversarial networks (GANs) to compute the distance between an example and its adversarial equivalent \cite{ducoffe_adversarial_2018}.

\subsection{Representativeness-based active learning}

One drawback of uncertainty-based methods is that they tend to select very similar images. In order to counter this problem, methods based on representativeness aim is to select a dataset $B*$ such that the images in this subset are as representative as possible of the whole dataset $B$.

A popular example of these technique is described in \cite{sener_active_2017}. The authors show that this problem can be solved with the geometric K-Center problem, using the L2 distance between activations of the last dense layer in the CNN. Using the CIFAR-10 and CIFAR-100 datasets, they compare their selection method with various other techniques, including a random selection, and the uncertainty approach by \cite{gal_deep_2017}. For both datasets, the technique seems to yield better results than all other approaches when the the labelling budget is less than 40\% of the dataset. According to the authors, the technique seems more efficient when the number of classes is low. They also conclude that uncertainty could be added to the recipe in order to improve the method.

A more recent work \cite{gissin_discriminative_2019} presents a technique inspired from GANs \cite{goodfellow_generative_2014} by deceiving a discriminator so that it can no longer decide if an example image selected by the generator belongs to $B*$ or $B$. Once the generator is trained well enough, it can be used to generate a dataset $B*$ which is representative of $B$. Using the MNIST and CIFAR-10 dataset, they compare it to other popular approaches and conclude that their method works as well as any other method. This approach is interesting because it is simple to adapt to other tasks such as object detection or segmentation. However, this technique requires an additional optimization step -- which makes the process much more computation-heavy than other methods.

\subsection{Hybrid active learning}

Few recent papers propose relevant methods that try to get the best of uncertainty and representativeness. Among them is \cite{ash_deep_2020}. The authors consider that uncertainty can be evaluated by studying the magnitude of the loss gradients in the last layer of the network: if the model is certain about a given prediction, the magnitude of the gradients would be low, and model parameters would only be weakly updated. The same loss gradients are used to evaluate diversity: a batch of examples has a good diversity if the gradients of the examples have complementary directions. Computing loss gradients is not possible if one does not have ground truth labels. To overcome this, the authors use model predictions instead of the groundtruth to compute the gradients. The authors test their method on a large diversity of use cases by varying the number of selected images, the CNN architecture used, and the dataset used (SVHN, OpenML \#156, CIFAR-10). They compare their method with the uncertainty-only approach and the diversity-only approach (Core-Set). The authors conclude that their approach is almost always the best, and that it seems to be robust to the diversity of use-cases.

\subsection{Other methods}

Some other methods in the litterature offer to solve the active learning problem, without directly measuring uncertainty or representativeness. This is the case of \cite{yoo_learning_2019}, where the authors propose to learn to predict the value of the loss during training by using a loss prediction module. The core hypothesis is that the loss value is directly linked to the quantity of information that a single example brings to the network during training. A clear advantage is that the module can be easily stacked to any architecture, meaning that the method can be easily adapted from the classification tasks to segmentation or detection tasks. For the detection task, the results seem to indicate that the Core-Set approach works better when the number of labelled images is low, whereas the loss prediction approach seems to work better when this number is higher.

Interestingly, some recent approaches have tried to treat the active learning problem as a reinforcement learning task. In \cite{casanova_reinforced_2020}, the authors propose to train a reinforcement learning algorithm to select the most useful patches in images from two public datasets -- CamVid and Cityscape -- to improve an existing segmentation model. Their solution is not straightforward to implement and remains more computation heavy than the others presented here but their results on both datasets seem to outperform other uncertainty-based approaches. It will be interesting to watch closely such reinforcement learning approaches in the future as they gain in maturity.

\section{Proposed approach}

In this work, we apply two active learning techniques to segmentation of satellite images: Bayesian dropout from \cite{gal_deep_2017} and Core-Set from \cite{sener_active_2017}. However, we need to adapt these methods to the segmentation task, as they were designed for image classification, and to allow the use of large rasters instead of pre-calibrated data.

\subsection{Pre-selection of tiles}

In order to use large raster images, the first step is to split the original images into fixed-size tiles. We choose 512x512 tiles to prevent any memory-related issues during training. However, considering the amount of data we use, processing all created tiles (several hundred thousands) through the active learning techniques proves unpractical. For this reason, we choose to perform a first selection on the tiles, before applying the active learning techniques. To do so, we use the initial network to predict once on all available tiles and we then order these tiles according to the average intensity of the corresponding segmentation map. This technique comes from initial experiments showing that tiles with a very low response were almost never among the ones selected by active learning techniques.

This pre-selection therefore allows to speed up considerably the whole process while not modifying the selection performed through active learning. The number of selected tiles is set to 5\% of all the available tiles, based on computational limitations.

\subsection{Bayesian dropout}

The first active learning method we use is derived from \cite{gal_deep_2017}. While this method was developed for image classification, we modify it to apply it for semantic segmentation. To do so, we follow the same approach of predicting multiple times on a given tile while activating dropout in the network, using the same parameters as the ones used during training. We choose to perform 10 predictions, to get sufficient variations while limiting the computing cost of the method.

We then need to derive an estimation of the uncertainty of the whole tile from the 10 segmentation maps thus obtained. We first compute, for each pixel, the variance of the 10 values. Then, we take the average of all the variances as an estimation of the uncertainty of the 512x512 tile.

\subsection{Core-Set}

The second active learning method we use uses the approach of \cite{sener_active_2017}. Similarly to the first method, this technique has been developed for image classification. To adapt it to segmentation, one need to find a way to derive a reasonably-sized vector from the features extracted by the network.

As the CNN we use has a U-Net structure, we use the feature map at the end of the decoder, which has a size of 128x128x128 (width, height, and number of filters). We first use max pooling to get an 8x8x128 matrix and then average pooling to compress it into a 1x1x128 matrix, interpreted as a 1D vector. We then use the Robust k-center algorithm from \cite{sener_active_2017} to select the k most representative tiles, k being equal to our labelling budget.

\section{Experiments and results}

To evaluate the potential of active learning for a defense-related use-case, we choose to test the developed methods on aircraft detection, using image segmentation models. As a first step, we performed only one iteration of the different methods and measure the performance increase compared with random selection.

\subsection{Initial models and pool of unlabelled images}

Before applying the active learning techniques, we need to train initial models which will be incrementally improved. We choose to create two different models, with the same modified U-Net architecture \cite{ronneberger_u-net_2015}, that correspond to two different use-cases:
\begin{itemize}
    \item a \emph{weak model} that has been trained on relatively few images, corresponding to a model still in development that we want to improve as fast as possible;
    \item a \emph{strong model} that has been trained on much more data and that reaches correct performances, representative of a model in production that could be improved with time.
\end{itemize}

We choose to select full-sized satellite images based on their date of acquisition, to mimic a real-life situation where images are gradually available. In this work, we use images from \emph{Maxar/DigitalGlobe} satellites WorldView~1, 2, and 3 and GeoEye~1. For the \emph{weak model}, we select images acquired 2010 January 1st and 2012 January 1st and for the \emph{strong model} we extend the second limit to 2017 January 1st. We then create a dataset from these images by splitting them into 512x512 tiles, with 90\% of the tiles being the ones containing aircraft and 10\% of the tiles being randomly sampled among negative tiles. We train each model until convergence, using the Adam optimizer \cite{kingma_adam_2014} and weighted cross-entropy loss. 

Now that we have initial models to improve, we need a vast amount of ``unlabelled'' images to select from. In our case, all the labels are already available, but only imagettes selected through active learning will be actually used. Following the same approach we followed to train the initial models, we select images from the \emph{Maxar/DigitalGlobe} satellites WorldView~1, 2, and 3 and GeoEye~1 based on their acquisition date. We therefore choose only images acquired between 2017 January 1st and 2017 July 1st. It is worth noting that while the images used to train the initial models contained mostly civilian aircraft, the images in our pool contains mostly military aircraft, which is an interesting setup to see if our models can improve on these new aircraft types.

Table~\ref{tab:initial_models} summarises the data used to train the initial models and to create the pool of ``unlabelled'' images.

\begin{table}[bt]
\centering
\caption{Data used to train the initial models and create the pool of unlabelled images.}\label{tab:initial_models}
\begin{tabular}{|c|c|c|c|c|c|}
\hline
Model & Acquisition period & \# images & Total area & \# aircraft & \# tiled imagettes\\
\hline
Weak & 2010/01/01-2012/01/01 & 40 & $\textrm{222~km}^\textrm{2}$ & 951 & 2,740 \\
Strong & 2010/01/01-2017/01/01 & 3,891 & $\textrm{14,445~km}^\textrm{2}$ & 67,008 & 54,851\\
\hline
Pool & 2017/01/01-2017/07/01 & 467 & $\textrm{5,905~km}^\textrm{2}$ & 11,017 & $\sim$~400k possible\\
\hline
\end{tabular}
\end{table}

\subsection{Testing set and performance metrics}

\begin{figure}[bt]
    \centering
    \includegraphics[width=12cm]{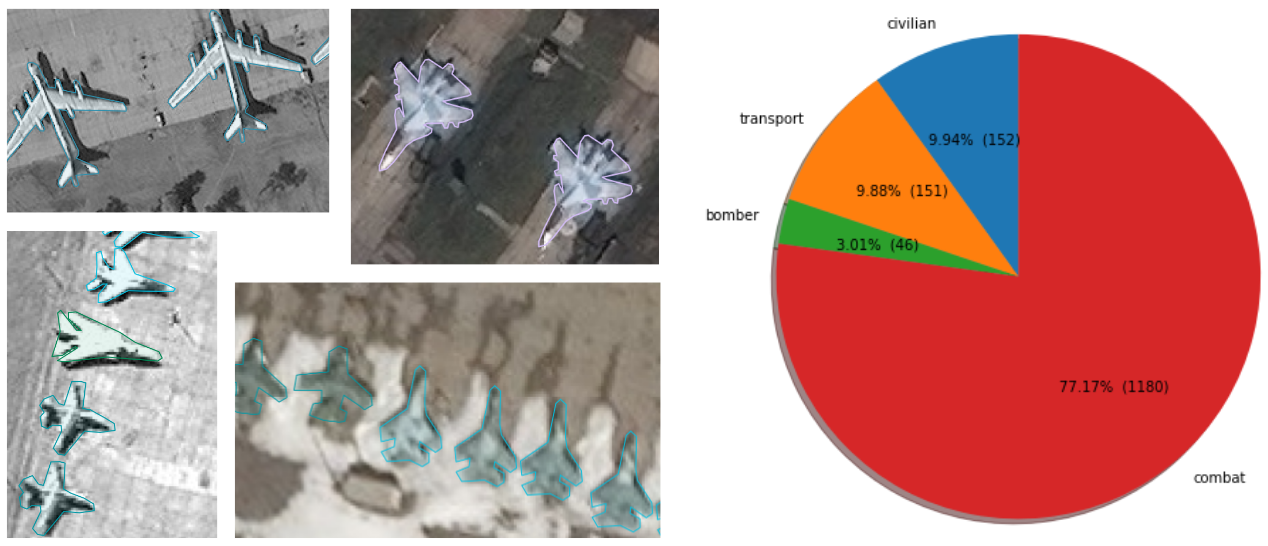}
    \caption{\label{fig:testing_set}%
    \textbf{Testing set used to measure performances. Left: examples of labelled aircraft. Right: Distribution of the aircraft models between four categories (combat, bomber, transport, and civilian).}}
\end{figure}

To measure the performances of the different models, we use a testing set composed of images present neither in the data used to train the initial models nor in the pool of unlabelled images. To focus the evaluation on military aircraft, we select 30 full-size satellite images from 16 different locations that contains mostly military aircraft. These images contain 1532 individual aircraft in total. Figure~\ref{fig:testing_set} presents examples of the labelled aircraft and the distribution of the models present in the testing set.

The metrics we consider are the classical precision, recall and F1-score. An aircraft in the ground truth is considered as a true positive if it is at least partly covered by the predicted segmentation. On the other hand, the segmentation outputed by the network is divided into connected components; if one component does not cover any aircraft in the ground truth, it is considered as a false positive. To compare the behaviour of different models, we consider the precision-recall curve of each model, obtained by varying the threshold applied to the segmentation output in order to obtain a binary map aircraft/background. We then choose as the operational threshold the one yielding the best F1-score.

\subsection{Results}

To evaluate the contribution of the uncertainty and core-set approaches, we compare them to two different baselines. The first baseline, called \emph{Unlimited}, corresponds to an unlimited labelling budget, with all the objects in the pool of images being labelled. In this case, a training set is built by selecting all the tiles containing aircraft (90\% of the set) and randomly selected negative tiles (10\% of the set), leading to a training set containing 5700 tiles, which is higher than the labelling budgets used for the active learning approaches. The second baseline, called \emph{Random}, uses the same labelling budget as the active learning approaches but in this case the tiles are randomly selected after the pre-selection step.

To improve the \emph{weak model}, we choose a labelling budget of 1000 tiles while for the \emph{strong model} we choose a higher budget of 5000 tiles. This choice was made to match the difference in the number of tiles in their initial training sets. Table~\ref{tab:budget} summarises the labelling budget and the number of tiles added to the initial datasets for the different approaches, as well as the gains in term of F1-score of each experiment.

\begin{table}[bt]
\centering
\caption{Labelling budget and number of tiles of the different approaches compared and gains in F1-score.}\label{tab:budget}
\begin{tabular}{|c|c|c|c|c|}
\hline
Model & Approach & Labelling budget & \# Tiles selected & F1-score gain \\
\hline
\multirow{4}{*}{Weak} & Baseline \emph{Unlimited} & unlimited & 5700 & +6.6\% \\
& Baseline \emph{Random} & 1000 & 1000 & +2.0\% \\
& Bayesian dropout & 1000 & 1000 & -0.2\% \\
& \textbf{Core-Set} & 1000 & 1000 & \textbf{+6.7\%} \\
\hline
\multirow{4}{*}{Strong} & Baseline \emph{Unlimited} & unlimited & 5700 & +3.6\% \\
& Baseline \emph{Random} & 5000 & 5000 & +4.5\% \\
& \textbf{Bayesian dropout} & 5000 & 5000 & \textbf{+8.0\%} \\
& Core-Set & 5000 & 5000 & +6.6\% \\
\hline
\end{tabular}
\end{table}

\begin{figure}[bt]
    \centering
    \includegraphics[width=6cm]{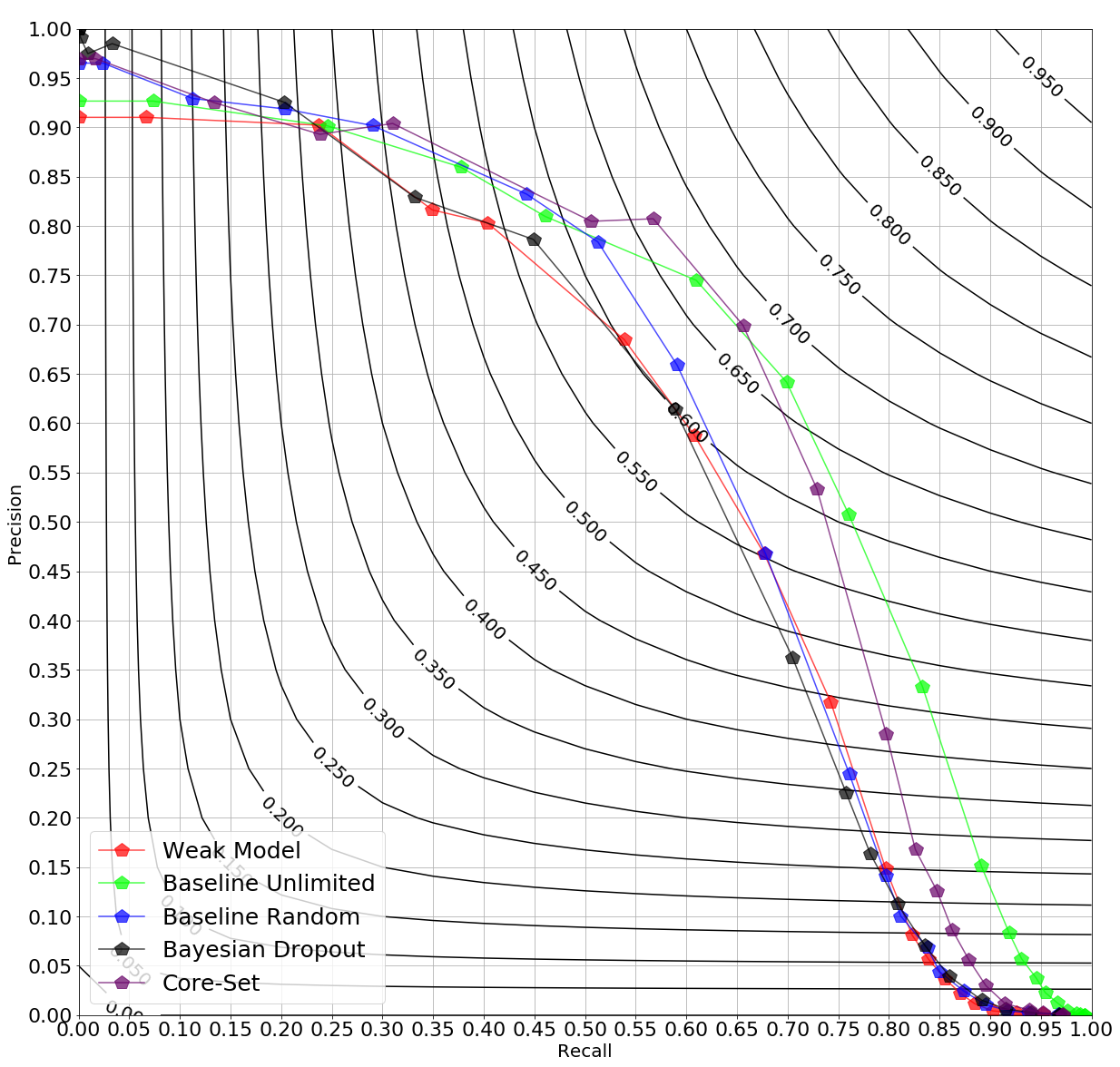}
    \includegraphics[width=6cm]{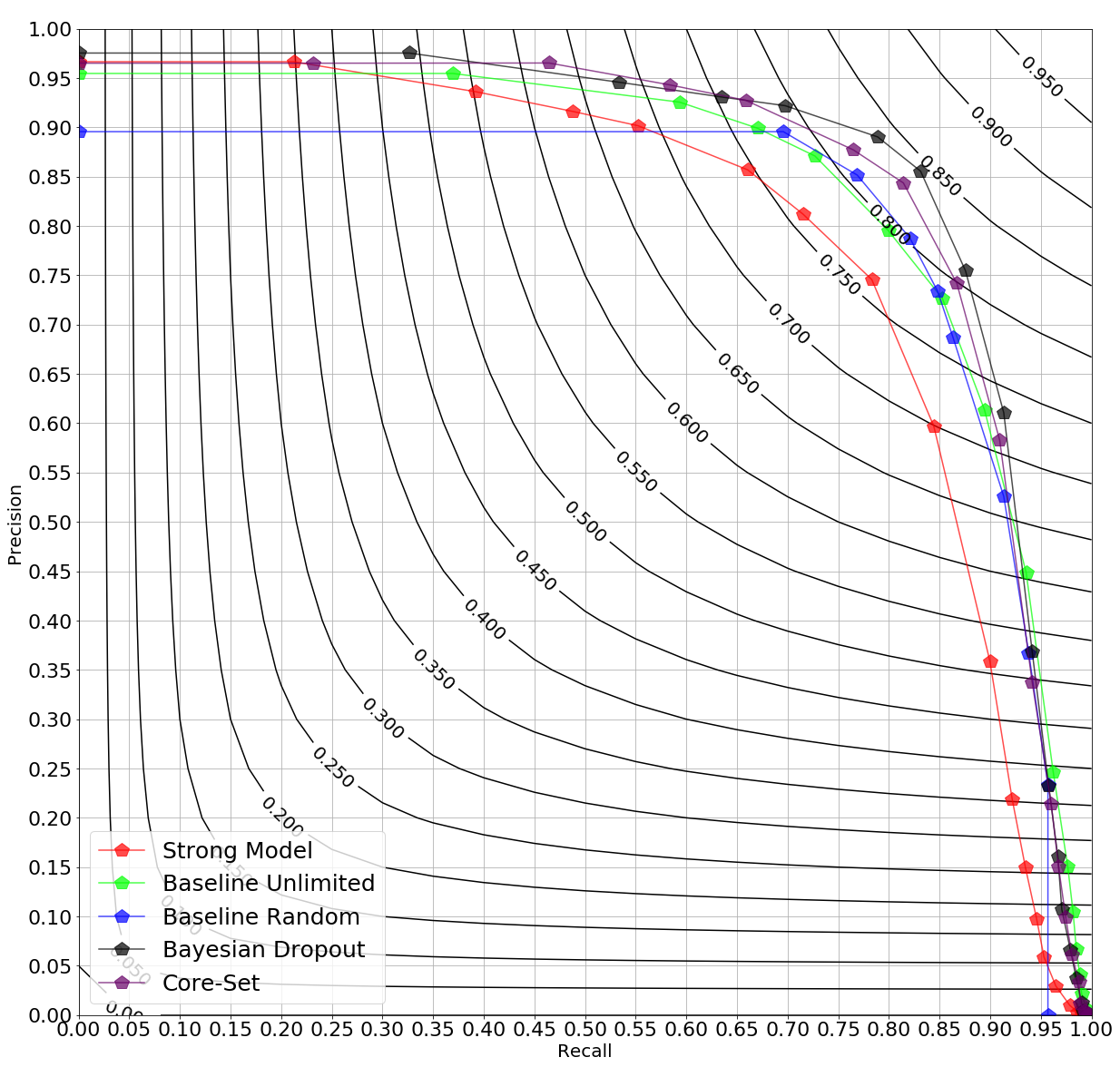}
    \caption{\label{fig:perfo}%
    \textbf{Performance of the various approaches. Left: weak model. Right: strong model. The contour lines indicate similar values of the F1-score.}}
\end{figure}

Results can be seen in Figure~\ref{fig:perfo}. For the \emph{weak model}, the best approach appears to be the Core-Set. It leads to an F1-score increase of 6.7\% while the Baseline \emph{Unlimited} -- that selects more than 5 times more tiles -- improves the F1-score of 6.6\%. However, in the high recall regime, the Baseline \emph{Unlimited} is performing better. This is likely due to the fact that the model trained for this baseline has bee presented with all the new aircraft while active learning techniques cannot necessarily select all the tiles containing aircraft due to their limited labelling budget. Interestingly, the Bayesian dropout approach does not lead to any significant improvement over the initial model and is outperformed by the Baseline \emph{Random}. This only exception lies in the very high precision regime, meaning that the Bayesian dropout impact is limited to removing some false positives detected by the initial \emph{weak model}.

For the \emph{strong model}, even if the Core-Set approach is also performing well (+6.6\% f F1-score), the Bayesian dropout approach outperforms it by a large margin (+8.0\% of F1-score). Both active learning approaches are behaving much better than the baselines -- including the \emph{Unlimited} one that selects 700 tiles more -- which shows the usefulness of selecting the examples to present to an already well-performing model to improve it. This follows the intuition than as a model gets more and more efficient, new data bring relatively small information to it. It is especially striking to see that active learning methods perform on part with the \emph{Unlimited} baseline in the high recall regime. This might be an indication that the \emph{strong model} has learned more generic features, which can be leveraged by both active learning methods to identify difficult aircraft models.

\section{Conclusion}

In this work, we have presented the various techniques of active learning that can be used to improve a object detection algorithm working on very high resolution remote sensing images. We have adapted two approaches -- Bayesian dropout and Core-Set -- to the segmentation task and tested them on two CNN models with different levels of performances. By comparing them with two baselines, we have shown that active learning approaches are very efficient to boost performances of existing models while limiting the amount of labels needed. The most consistent active learning technique tested here seems to be the Core-Set one, as it yields good results on both test cases. However, the Bayesian dropout technique is significantly more efficient for the strong model, which could imply the necessity to change the active learning strategy at one stage during model development.

Future developments could include a variety of different experiments. First, it could be informative to apply these two techniques in an iterative manner, with several pools of data. It could allow to identify at which point the uncertainty approach becomes more efficient than the core-set one. Another obvious option would be to follow the same experimental on other types objects, such as terrestrial vehicles or vessels for instance. Finally, these techniques still rely on the initial training set, which both makes the new trainings long and can be impractical in an industrial context where data cannot be transferred. Combining the active learning techniques developed in this paper with continuous learning such as Elastic Weight Consolidation \cite{kirkpatrick_overcoming_2017} to avoid re-using initial data could lead to an even more operational solution.

Even with the aforementioned cautions, we believe that this work has proven how efficient and relevant active learning solutions are in a defense-related context. The techniques presented here have a high potential to ease adoption of CNN-based solutions into the workflow of military analysts. They can allow them to adapt products developed on commercial data to their own specific needs, either linked to sovereign sensors or to new types of objects of interest.

%
%
%
\bibliographystyle{splncs04}
\bibliography{references}

\begin{thebibliography}{10}
\providecommand{\url}[1]{\texttt{#1}}
\providecommand{\urlprefix}{URL }
\providecommand{\doi}[1]{https://doi.org/#1}

\bibitem{ash_deep_2020}
Ash, J.T., Zhang, C., Krishnamurthy, A., Langford, J., Agarwal, A.: Deep
  {Batch} {Active} {Learning} by {Diverse}, {Uncertain} {Gradient} {Lower}
  {Bounds}. In: {ICLR} (2020)

\bibitem{beluch_power_2018}
Beluch, W.H., Genewein, T., Nürnberger, A., Köhler, J.M.: The power of
  ensembles for active learning in image classification. In: Proceedings of the
  {IEEE} {Conference} on {Computer} {Vision} and {Pattern} {Recognition}. pp.
  9368--9377 (2018)

\bibitem{budd_survey_2019}
Budd, S., Robinson, E.C., Kainz, B.: A {Survey} on {Active} {Learning} and
  {Human}-in-the-{Loop} {Deep} {Learning} for {Medical} {Image} {Analysis}.
  arXiv preprint arXiv:1910.02923  (2019)

\bibitem{casanova_reinforced_2020}
Casanova, A., Pinheiro, P.O., Rostamzadeh, N., Pal, C.J.: Reinforced active
  learning for image segmentation. arXiv preprint arXiv:2002.06583  (2020)

\bibitem{ducoffe_adversarial_2018}
Ducoffe, M., Precioso, F.: Adversarial active learning for deep networks: a
  margin based approach. arXiv preprint arXiv:1802.09841  (2018)

\bibitem{gal_deep_2017}
Gal, Y., Islam, R., Ghahramani, Z.: Deep bayesian active learning with image
  data. arXiv preprint arXiv:1703.02910  (2017)

\bibitem{gissin_discriminative_2019}
Gissin, D., Shalev-Shwartz, S.: Discriminative active learning. arXiv preprint
  arXiv:1907.06347  (2019)

\bibitem{goodfellow_generative_2014}
Goodfellow, I., Pouget-Abadie, J., Mirza, M., Xu, B., Warde-Farley, D., Ozair,
  S., Courville, A., Bengio, Y.: Generative adversarial nets. In: Advances in
  neural information processing systems. pp. 2672--2680 (2014),
  \url{http://arxiv.org/abs/1406.2661}, arXiv: 1406.2661

\bibitem{kingma_adam_2014}
Kingma, D.P., Ba, J.: Adam: {A} method for stochastic optimization. arXiv
  preprint arXiv:1412.6980  (2014)

\bibitem{kirkpatrick_overcoming_2017}
Kirkpatrick, J., Pascanu, R., Rabinowitz, N., Veness, J., Desjardins, G., Rusu,
  A.A., Milan, K., Quan, J., Ramalho, T., Grabska-Barwinska, A., {others}:
  Overcoming catastrophic forgetting in neural networks. Proceedings of the
  national academy of sciences  \textbf{114}(13),  3521--3526 (2017),
  publisher: National Acad Sciences

\bibitem{ronneberger_u-net_2015}
Ronneberger, O., Fischer, P., Brox, T.: U-net: {Convolutional} networks for
  biomedical image segmentation. In: International {Conference} on {Medical}
  image computing and computer-assisted intervention. pp. 234--241. Springer
  (2015)

\bibitem{sener_active_2017}
Sener, O., Savarese, S.: Active learning for convolutional neural networks: {A}
  core-set approach. arXiv preprint arXiv:1708.00489  (2017)

\bibitem{yoo_learning_2019}
Yoo, D., Kweon, I.S.: Learning loss for active learning. In: Proceedings of the
  {IEEE} {Conference} on {Computer} {Vision} and {Pattern} {Recognition}. pp.
  93--102 (2019)

\end{thebibliography}

\end{document}